\documentclass[final]{cvpr}

\usepackage{times}
\usepackage{epsfig}
\usepackage{graphicx}
\usepackage{amsmath}
\usepackage{amssymb}

\usepackage{array} 
\usepackage{booktabs}
\usepackage{microtype}

\newcommand*{\textBF}[1]{\rlap{\raisebox{0pt}[0pt][0pt]{\textbf{#1}}}\phantom{#1}}

\usepackage[pagebackref=true,breaklinks=true,colorlinks,bookmarks=false]{hyperref}

\def\cvprPaperID{4} 
\def\confName{CVPR}
\def\confYear{2021}

\begin{document}

\title{OmniFlow: Human Omnidirectional Optical Flow}

\author{Roman Seidel, André Apitzsch, Gangolf Hirtz\\
Chemnitz University of Technology\\
Faculty of Electrical Engineering and Information Technology\\
09126 Chemnitz, Germany\\
{\tt\small \{roman.seidel,andre.apitzsch,g.hirtz\}@etit.tu-chemnitz.de}
}

\maketitle

\begin{abstract}
Optical flow is the motion of a pixel between at least two consecutive video frames and can be estimated through an end-to-end trainable convolutional neural network.
To this end, large training datasets are required to improve the accuracy of optical flow estimation. 
Our paper presents OmniFlow: a new synthetic omnidirectional human optical flow dataset.
Based on a rendering engine we create a naturalistic 3D indoor environment with textured rooms, characters, actions, objects, illumination and motion blur where all components of the environment are shuffled during the data capturing process.
The simulation has as output rendered images of household activities and the corresponding forward and backward optical flow.
To verify the data for training volumetric correspondence networks for optical flow estimation we train different subsets of the data and test on OmniFlow with and without Test-Time-Augmentation.
As a result we have generated 23,653 image pairs and corresponding forward and backward optical flow.
Our dataset can be downloaded from: \href{https://www.tu-chemnitz.de/etit/dst/forschung/comp_vision/datasets/omniflow/}{\texttt{https://mytuc.org/byfs}}
\end{abstract}


\section{Introduction}

In the last decade, large-scale synthetic training and benchmark datasets have driven innovation in computer vision, have accelerated the development of learning-based approaches and demonstrated a way of quantitative evaluation without capturing real-world data.
Especially in optical flow estimation this plays an important role due to the high necessary effort to collect and label real-world data. 
Optical flow is the motion of a pixel between at least two consecutive video frames and can be estimated through an end-to-end trainable convolutional neural network (CNN) \cite{dosovitskiy2015flownet, ilg_flownet_2017}.


Assuming an indoor scenario with fisheye cameras with a field of view (FOV) of greater than or equal to 180$^\circ$ a whole room can be captured with only one sensor.
Fisheye cameras follow the omnidirectional camera model and the question arises whether to formulate their distortions implicitly in the model architecture or explicitly generate synthetic omnidirectional data.

Perspective human optical flow cannot be used to determine human motions in omnidirectional images due to unsuitable layer architectures or missing data.
However, our data-driven approach generates a dataset with human optical flow in omnidirectional images that contains distortions and invariances against angle of view.
Our dataset contains various indoor household activities such as \textit{sitting down and standing up}, \textit{walking} or \textit{falling down}.
Human optical flow for omnidirectional images can be used for computer vision tasks such as motion estimation, indoor navigation of robots and tracking of persons with indoor surveillance systems.


\section{Related Work}

%


%
Synthetic optical flow datasets for training or benchmarking CNNs are extensively available and firstly initiated by the Middlebury\cite{baker2011database} and the MPI Sintel dataset\cite{butler2012naturalistic}.
In parallel, with a new correlation-based network architecture datasets with synthetic foreground and real-world random background images were created by~\cite{dosovitskiy2015flownet, ilg_flownet_2017}, namely FlyingChairs and FlyingThings. 
Aodha \etal~\cite{mac_aodha_segmenting_2010} created a system for easily producing synthetic ground truth data for both optical flow and descriptor-matching, where the pipeline lacks realistic scenarios and has no humans included.
With the goal of benchmarking algorithms in the perception of the environment of autonomous cars the KITTI Benchmark Suite~\cite{geiger2012we} and -- for multi object tracking --
VirtualKITTI~\cite{gaidon2016virtual, cabon_virtual_2020} were created.
Both have optical flow ground truth; KITTI from a lidar sensor and virtual KITTI through a 3D rendering engine.
A tabular overview of optical flow training and benchmark datasets based on perspective views is shown in \autoref{tab:compflow}.



\begin{table*}[ht!]
\begin{center}
\resizebox{0.93\linewidth}{!}{%
\begin{tabular}{l>{\centering}m{1.5cm}>{\centering}m{1.8cm}rr>{\centering}m{2.5cm}>{\centering}m{2cm}}
\toprule
Dataset        & Synthetic/ Natural & Training/ Benchmark & \#Frames & Resolution & Moving/Static camera & Person/ Non-person \tabularnewline
\midrule
Driving         \cite{mayer2016large}          & S   & T   & \(4{,}392\)            & \(960 \times 540\)         & M                 & N    \tabularnewline
FlyingChairs    \cite{dosovitskiy2015flownet}  & S   & T   & \(21{,}818\)           & \(512 \times 384\)         & M                 & N    \tabularnewline
FlyingThings3D  \cite{mayer2016large}          & S   & T   & \(22{,}872\)           & \(960 \times 540\)         & M                 & N    \tabularnewline
HD1K            \cite{kondermann2016hci}       & N   & B   & \(1{,}083\)            & \(2{,}560 \times 1{,}080\) & M                 & P    \tabularnewline
KITTI 2012      \cite{geiger2012we}            & N   & B   & \(194\)                & \(1{,}242 \times 375\)     & M                 & (P)  \tabularnewline
KITTI 2015      \cite{menze2015object}         & N   & B   & \(200\)                & \(1{,}242 \times 375\)     & M                 & (P)  \tabularnewline
Middlebury      \cite{baker2011database}       & S/N & B   & \(8\)                  & \(640 \times 480\)         & SC                & N    \tabularnewline
Monkaa          \cite{mayer2016large}          & S   & T   & \(8{,}591\)            & \(960 \times 540\)         & M                 & N    \tabularnewline
SceneNet RGB-D  \cite{mccormac2017scenenet}    & S   & T   & \(\sim 5{,}000{,}000\) & \(320 \times 240\)         & M                 & N    \tabularnewline
Sintel          \cite{butler2012naturalistic}  & S   & B/T & \(1{,}064\)            & \(1{,}024 \times 436\)     & M                 & (P)  \tabularnewline
SplitSphere     \cite{huguet_variational_2007} & S   & B   & unknown                & \(512 \times 512\)         & SC                & N    \tabularnewline
UCL             \cite{mac2010segmenting}       & S   & B   & \(4\)                  & \(640 \times 480\)         & SC                & N    \tabularnewline
UCL (extended)  \cite{mac2012learning}         & S   & B   & \(20\)                 & \(640 \times 480\)         & SC                & N    \tabularnewline
Virtual KITTI   \cite{gaidon2016virtual}       & S   & B/T & \(21{,}260\)           & \(1{,}242 \times 375\)     & M                 & P    \tabularnewline
Virtual KITTI 2 \cite{cabon_virtual_2020}      & S   & B/T & \(21{,}260\)+\(2{,}126\) & \(1{,}242 \times 375\)   & M                 & P    \tabularnewline
\textBF{OmniFlow (ours)}                       & S   & B/T & \(23{,}653\)           & {\(2{,}048 \times 2{,}048\)}     & Static per scene  & P    \tabularnewline
\bottomrule
\end{tabular}
}
\end{center}
\caption{Comparison of optical flow benchmark and training datasets.}
\label{tab:compflow}
\end{table*}

Single and multi-human optical flow datasets in perspective front views were investigated in~\cite{ranjan_learning_2018, ranjan_learning_2020} from which we differ in terms of camera geometry, body models and output image and flow resolution. 
A dataset which focuses on the application of crowd analysis was created in~\cite{schroder_optical_2018} where an omnidirectional synthetic dataset which contains bounding boxes, segmentation masks and depth maps in indoor scenarios~\cite{scheck_learning_2020} lacks optical flow.


\section{OmniFlow: Human Omnidirectional Optical Flow}

\begin{figure*}[ht]
\begin{center}
\includegraphics[width=0.8\linewidth]{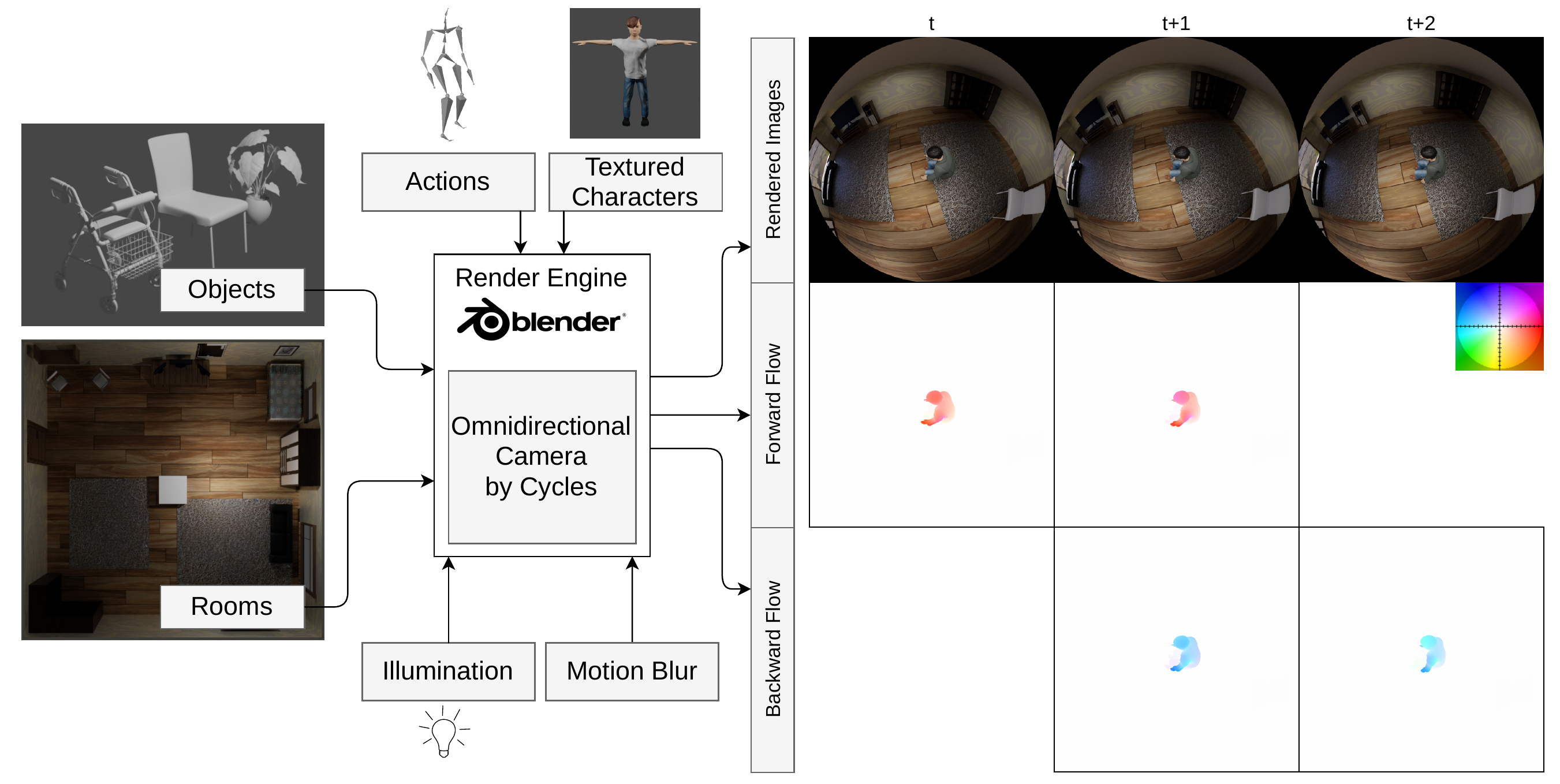}
\end{center}
\caption{OmniFlow dataset creation pipeline. We rig textured characters to actions, place both in randomly selected rooms and add objects, illumination and (to some scenes) motion blur to generate high resolution image sequences with corresponding forward and backward flow.}
\label{fig:datapipeline}
\end{figure*}

This section describes the dataset creation pipeline of human omnidirectional optical flow.
Our data is created by the rendering engine \textit{Blender}, which we used to model a 3D indoor environment with randomly placed animated humans in various rooms, objects and a virtual camera with an omnidirectional camera geometry.
The whole dataset creation pipeline is shown in \autoref{fig:datapipeline}.
The setup of these animated scenes contains a moving human from various camera locations with static background modeled as 3D indoor environment.
We have generated 321 scenes each with 75 images and corresponding forward and backward flow and a total dataset size of 23,653 frames containing 18,921 frames for training and 2,366 frames each for validation and testing.

{\bf Dataset Creation Pipeline.}
To counter the gap between synthetic and real-world image data domain randomization \cite{tremblay_training_2018} is implemented in our simulation.
To this end, randomly set simulation parameters affect the building process of the entire scene.
Various skeletons were used by the simulation separately for each scene while person models are rigged to these skeletons.
Further simulation parameters are rooms with different textures and random illumination size and energy from two area lights.
Additionally, objects such as tables, chairs or plants are placed inside each room.
To achieve various viewing angles of the camera with respect to the character the camera location is variable, where the extent of the room is used for potential camera viewpoints.

{\bf Rendering Engine.}
The optical flow of OmniFlow is created by \textit{vector pass} of Blender's Cycles rendering engine which exclusively delivers an omnidirectional camera, as shown in \autoref{fig:datapipeline}.
Given three consecutive image frames \(t\), \(t+1\) and \(t+2\) the \textit{vector pass} contains motion between these frames in forward (images \(t\) and \(t+1\)) and backward direction (images \(t+1\) and \(t+2\)).

{\bf Camera.}
For rendering the 3D environment to 2D images, a virtual camera with the fisheye-equidistant camera model is randomly placed in a 4\(\times\)4 m square in the center of the room.
The camera is tracked to the hip bone of the animation to make sure that the human is centered with respect to the camera image.
To cover the whole scene, the FOV is 180$^{\circ}$ without considering the sensor type and sensor size.

{\bf Rooms.}
On the basis of \cite{scheck_learning_2020} we use four different rooms from ProBuilder, a world building tool in the Unity Editor with random textures\footnote{\url{https://www.cc0textures.com}} which are changed with every scene.
We import the geometries and the corresponding textures for walls, ground and furniture to Blender.
The rooms itself have a spatial expansion of 20\(\times\)20 m and a height of 4 m.

{\bf Animations.}
We include various animations from the Motionbuilder-friendly BVH conversion release of CMU's Motion Capture Database\footnote{\url{http://mocap.cs.cmu.edu/} and \url{http://cgspeed.com}} where frame-1 T-poses are included.
Examples of animations from the CMU Motion Capture database are activities of the humans, e.g. \textit{sitting down and standing up, falling, tooth brushing, brooming} and can be adapted to every other daily activity. %

{\bf Characters.}
The characters were generated by \textit{MakeHuman} and contain 16 user generated textures with upper body clothes, lower body clothes, shoes and body extensions (e.g. hairstyles).
All characters were set to T-pose (base animation pose), rigged to CMU's hybrid dataset and randomized in terms of weight, height, age and gender.

{\bf Objects.}
Natural objects, such as potted plants, chairs and tables are included in the scene to make the indoor environment realistic and evoke occlusions of objects with the animated humans.

{\bf Illumination.}
For indoor lighting we randomly set two light sources in each scene.
An area-light directly above the character and a ceiling light with a distance of \(2.5\) to \(7.5\) meters to the character to get a realistic lightning scenario.
The light varies for each scene in terms of energy and position in the above specified limits.
Naturalistic outdoor illumination was realized by High Dynamic Range Images (HDRI) from HDRIHeaven\footnote{\url{https://hdrihaven.com}}.
With realistic day, dusk, dawn and night HDRIs we were able to simulate different times of day.
Both, indoor lights and outdoor illumination were selected randomly in our scenes.

{\bf Motion Blur.}
Since fast-moving objects are appearing not constantly sharp during their motion, we applied motion blur to the human activity and switch it randomly on and off in each scene.


\section{Evaluation}

\begin{figure}[t]
\begin{center}
   \includegraphics[width=0.84\linewidth]{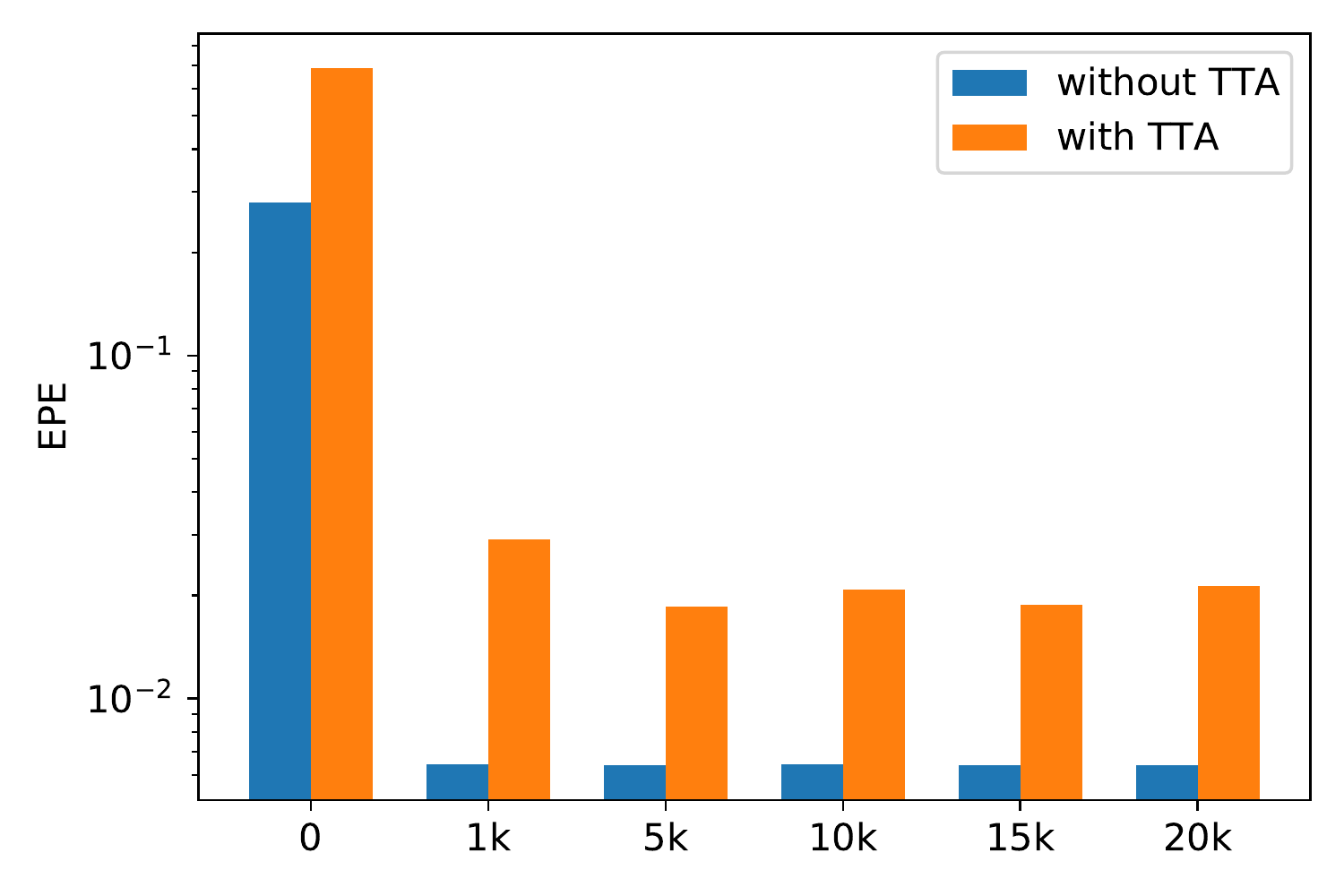}
\end{center}
   \caption{Determination of a \textit{sufficient} amount of data for training RAFT.
    0 means that we test a on FlyingChairs and FlyingThings pretrained model on test split of OmniFlow.
    Furthermore, we fine-tune RAFT on our data on selected subsets of 1k, 5k, 10k, 15k and 20k data with and without test-time augmentation.}
\label{fig:num_images}
\end{figure}

Our dataset is evaluated on a test set of OmniFlow of 10\% of the whole dataset.
We train a correspondence network for optical flow and fine-tune on five subsets 1k, 5k, 10k, 15k and 20k of OmniFlow with a pretrained model on FlyingChairs and FlyingThings.
As long there is no further omnidirectional optical flow dataset for testing available we use test-time augmentation (TTA) \cite{shanmugam_when_2020} with three standard augmentation methods cropping, scaling and horizonal flipping.
Results on OmniFlow test set are shown in \autoref{fig:num_images}.

{\bf Recurrent All Pairs Field Transforms.} Recurrent All Pairs Field Transforms for Optical Flow (RAFT) consists of a per-pixel feature encoder for extracting features from both input images \(I_1, I_2\) and a context encoder that extracts features only from \(I_1\).
A correlation layer generates a 4D correlation tensor for all pairs of pixels including subsequent pooling to produce lower resolution volumes.
The update operator based on a gated activation unit (GRU-block) where fully connected layers are replaced by convolutions.
The input feature map of the update operator is the concatenation of correlation, flow and context features.
In general, we follow the training schedule of RAFT but resize our OmniFlow data to a input resolution of 512\(\times\)512 px and use a learning rate of 1e-6 and a weight decay of 1e-5.

{\bf Our Observations.} 
We figure out that 5,000 image pairs of OmniFlow are sufficient to fine-tune RAFT with synthetic omnidirectional human optical flow data.
Nevertheless we provide approx. 20,000 image pairs where CNNs based on correlation layers needs more data for training.



\section{Conclusion}

In this paper, we create OmniFlow: a new omnidirectional human optical flow dataset.
With a 3D rendering engine, namely \textit{Blender} we generate a naturalistic 3D indoor environment with textured rooms, characters, actions, objects, illumination and motion blur.
We evaluate our data with TTA and have explored that the amount of 5k images is sufficient to fine-tune RAFT on FlyingChairs and FlyingThings.
Our next steps are the investigation in other network architectures for correlation-based or semi-supervised optical flow CNNs and the interpretation of optical flow as fine-grained human activities.



{\small
\bibliographystyle{ieee_fullname}
\bibliography{2021-flow-literature}
}

\end{document}